\documentclass[lettersize,journal]{IEEEtran}
\usepackage{amsmath,amsfonts}
\usepackage{algorithmic}
\usepackage{algorithm}
\usepackage{array}
\usepackage[caption=false,font=normalsize,labelfont=sf,textfont=sf]{subfig}
\usepackage{textcomp}
\usepackage{stfloats}
\usepackage{url}
\usepackage{verbatim}
\usepackage{graphicx}
\usepackage{cite}
\usepackage{multicol}
\usepackage{xcolor}

\begin{document}

\title{Soft Sensing Regression Model: from Sensor to Wafer Metrology Forecasting}

\author{Angzhi Fan \IEEEauthorrefmark{1} \IEEEauthorrefmark{2}, Yu Huang  \IEEEauthorrefmark{1}, Fei Xu \IEEEauthorrefmark{1} \IEEEauthorrefmark{2}, Sthitie Bom \IEEEauthorrefmark{1}

\begin{multicols}{2}
\IEEEauthorblockA{\IEEEauthorrefmark{1} Seagate Technology, MN, USA 
    \\\{yu.1.huang, sthitie.e.bom\}@seagate.com} \\\
\IEEEauthorblockA{\IEEEauthorrefmark{2} University of Chicago, IL, USA 
    \\\{fana, feixu\}@uchicago.edu}
\end{multicols}
\thanks{This work was produced when the authors worked in the Global Wafer Systems organization at Seagate Technology from June to September, 2022.}
\thanks{Manuscript received Jan XX, 2023; revised Jan XX, 2023.}
}

\markboth{Journal of \LaTeX\ Class Files,~Vol.~14, No.~8, August~2021}%
{Shell \MakeLowercase{\textit{et al.}}: A Sample Article Using IEEEtran.cls for IEEE Journals}



\maketitle

\begin{abstract}
The semiconductor industry is one of the most technology-evolving and capital-intensive market sectors. Effective inspection and metrology are necessary to improve product yield, increase product quality and reduce costs. In recent years, many semiconductor manufacturing equipments are equipped with sensors to facilitate real-time monitoring of the production process. These production-state and equipment-state sensor data provide an opportunity to practice machine-learning technologies in various domains, such as anomaly/fault detection, maintenance scheduling, quality prediction, etc. In this work, we focus on the task of soft sensing regression, which uses sensor data to predict impending inspection measurements that used to be measured in wafer inspection and metrology systems. We proposed an LSTM-based regressor and designed two loss functions for model training. Although engineers may look at our prediction errors in a subjective manner, a new piece-wise evaluation metric was proposed for assessing model accuracy in a mathematical way. The experimental results demonstrated that the proposed model can achieve accurate and early prediction of various types of inspections in complicated manufacturing processes.
\end{abstract}


\section{Introduction}
\IEEEPARstart{T}{he} development of AI technologies is of strategic significance to the
semiconductor industry because of their potential for reducing the capital footprint and improving cycle time and yields.
There is a significant opportunity to innovate novel machine learning techniques to make our manufacturing processes and controls smarter and more efficient. Soft sensing, being one of those emerging techniques, plays an important role in monitoring industrial processes. In general terms, soft sensing models can be defined as inferential models that use \textit{easy-to-measure} variables (e.g. online available sensors) for online estimation of \textit{hard-to-measure} variables (e.g. quality variables). However, the development of AI-based soft sensing models is extremely sluggish and erratic. The major concern is that most semiconductor manufacturing systems are
customized for specific applications with limited scalability. The AI model development in semiconductor industries suffered from poor flexibility and always involved extensive
prior knowledge of semiconductor manufacturing mechanisms. In this paper, we aim at developing a purely data-driven machine-learning-based soft sensing model for regression applications, specifically for semiconductor metrology, providing a practicable solution for semiconductor industries. 

Soft sensing for regression applications are predictive models that use online available sensors recordings (e.g. pressure, voltage, etc.) to predict quality indicators that cannot be automatically measured at all,
or can only be measured at the follow-up metrology stage which is time-consuming, high cost, or sporadically. During the wafer manufacturing process at Seagate factories, there are a lot of sensor data stored from the processes. This is the basis for the subsequent use of such data for the development of related data-driven soft-sensing models. The semiconductor manufacturing process is becoming increasingly more complex and longer, resulting in a significant increase in cost and difficulty in measuring the key quality indicators. After a wafer is manufactured, the engineers use metrology tools to do the time-consuming laboratory analysis. It is highly desirable to develop a soft-sensing system for defective wafer detection, to save the time and capacity those metrology tools. 

There are two categories of soft-sensing models, \textit{white-box} physical models and \textit{black-box} data-driven models. Modeling by physical approach requires prior solid knowledge and usually focuses on the ideal steady-state processing, which is not feasible for complex systems. The data-driven models are based on historical observations of the process and are able to predict the real conditions of the process while physical models are unable to do. According to the sensor data collected in Seagate factories, it's impossible to have a physical model for regression application in metrology. Therefore, the remaining difficulties are related to the difficulty of choosing the correct model type and
structure necessary for the development.

There are several difficulties in soft sensing regression. First, the dependencies between sensor readings and measurements might be very complicated. Second, there is variability in the  engineering decision of whether a wafer is pass or fail. Third, there are a lot of missing data and even measurement errors in the dataset. Due to the above characteristics in data, a nonlinear data-driven model should be the best choice.

Our approach to dealing with soft sensing regression is based on deep learning methods. For a wafer, its sensor readings include one or multiple time steps, and we call them \emph{sensor time steps}. More details are explained in section \ref{sec:data}. Deep learning models, such as Long Short-term Memory (LSTM) \cite{lstm} network, GRU \cite{GRU}, and Transformer  \cite{transformer}, are popular tools to tackle sequential data. The recurrent structure in LSTM \cite{lstm} is capable of handling sequential data of different lengths. The input gate, output gate and forget gate in LSTM \cite{lstm} prevent the vanishing gradient problem to some degree. LSTM-based models are widely applied in Natural
Language Processing (NLP) and related tasks, such as text classification \cite{textclassification}, machine translation \cite{machinetranslation} and speech recognition \cite{speech}. Since our sensor time steps are also sequential data like sentences in NLP, it's a natural idea to apply these NLP tools into our task. A sensor time step corresponds to a word in the sentence. 

The main contributions of this work include the following: First, we present the wafer soft sensing regression dataset in detail, and provide a way of understanding and preprocessing it. Second, we design two loss functions to handle the different precision requirements of the wafers.
Third, we formulate the somewhat subjective evaluation criteria by engineers into several mathematical evaluation criteria.

The rest of this article is organized as follows. Section \ref{sec:relatedwork} summarizes some existing works related to our soft sensing regression problem. Section \ref{sec:method} introduces our model architecture and loss functions. Our data and our data preprocessing method are explained in section \ref{sec:data}. Experiments are shown in section \ref{sec:results}. Finally, we have our discussions and conclusion in section \ref{sec:discussion_conclusion}.

\section{Related Work}
\label{sec:relatedwork}
In the field of soft sensing, a recent survey paper \cite{survey_softsensing} summarizes many deep learning techniques. This paper emphasizes the use of four techniques: Autoencoder (AE) \cite{AEKramer1991, VAEKingma2014AutoEncodingVB}, Restricted Boltzmann Machine (RBM) \cite{RBMsmolensky1986information, RBMHinton2012APG}, Convolutional Neural Network (CNN) \cite{goodfellow2016deep} and Recurrent Neural Network (RNN) \cite{RNNrumelhart1986learning}. Among these four techniques, RNN and its variants are very closely related to our soft sensing regression task. 

RNN-based or LSTM-based soft sensors have a lot of applications on soft sensing data which has strong sequential characteristics, like chemical industrial processes \cite{LSTM-chemical} and the contact area between tires and the ground \cite{tires}. Researchers have also made some attempts to modify the model architectures. LSTM-FCN \cite{LSTM-FCN} combines fully convolutional neural networks (FCN) with LSTM to improve the performance. SLSTM \cite{SLSTM}, DLSTM \cite{DLSTM} and SBiLSTM \cite{SBiLSTM} utilize quality information in the modeling. Our task is different in that we may have various types of measurements for each wafer and we may want to train them jointly. These measurement types are characterized by several categorical variables. And because of this, soft sensing regression needs to include those categorical variables as input.

One task similar to ours is soft sensing classification. Soft Sensing Transformer (SST) \cite{SoftSensingTransformer} demonstrates the similarities between sensor readings and text data, and applies Transformer encoder \cite{transformer} into this task. ConFormer \cite{ConFormer} integrates the structures of both CNN and Transformer. Soft Sensing Model Visualization \cite{visualization} put more focus on model interpretation. GraSSNet \cite{GraSSNet} is a flexible Graph Neural Network (GNN) \cite{GNN, GCN} model applied on soft sensing classification. 

There are two big differences between soft sensing regression and classification. First, the class imbalanced issue can be a major issue bothering soft sensing classification, but a regression task does not suffer that much from it. Second, soft sensing regression can tell us how far the measurement deviates from acceptable values, but soft sensing classification only tells us the wafer is pass or fail. 

\section{Method}
\label{sec:method}
The inputs to our model consist of sensor features and measurement features, where measurement features are referred to as the categorical variables in measurements. A wafer may be measured at multiple locations, and the median of those measurement results is called \emph{meas\_med}. In this work our target variable is \emph{meas\_med}, and the output of our model is trying to predict our target.

To deal with sequential data like sensor time steps, we choose LSTM to build our model as explained in section \ref{subsec:models}.

Loss function is another important factor in this task. In section \ref{subsec:losses}, we introduce two loss functions for our soft sensing regression task. The usual $L_2$ does not make much sense in our task because different wafers usually have different precision requirements. 

\subsection{Model architecture}
\label{subsec:models}
\begin{figure}[!t]
  \centering
  \includegraphics[width=2.5in]{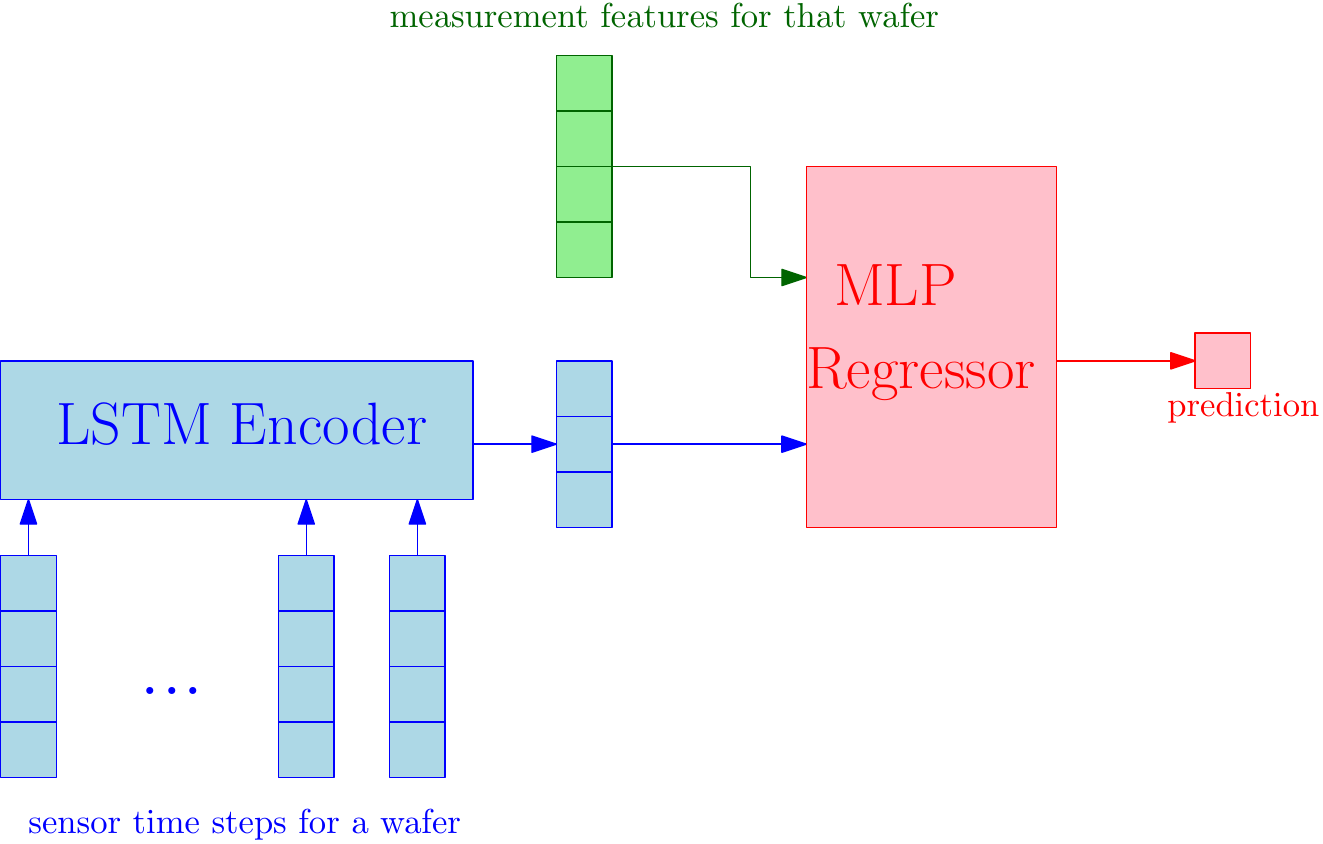}
  \caption{LSTM model.}
  \label{fig:LSTM}
\end{figure}

As shown in Figure \ref{fig:LSTM}, for a uniquely identified wafer, we use an encoder to encode its sensor time steps into a fixed-dimensional sensor vector. Our sensor encoder is a one-layer LSTM encoder with an embedding layer, and our encoded sensor vector is the last cell state of the LSTM encoder. The embedding layer between the sensor time steps and the LSTM structure can help dealing with the categorical variables in the sensor time steps. 

For every one of the measurements from the same wafer, we concatenate the measurement features with the encoded sensor vector and use an Multiple Layer Perceptron (MLP) to predict the target.

This procedure is consistent with the chronological order of the wafer manufacturing process, because a wafer first goes through the machines where we get sensor readings and then gets measured by metrology tools. 

\subsection{Loss functions}
\label{subsec:losses}

\subsubsection{Relative Error Loss}
For ground truth $y$ and prediction $\hat{y}$, the  \emph{Relative Error} is defined as 
\begin{equation}
  \eta =\frac{|\hat{y}-y|}{|y|}
  \label{eq:RE}
\end{equation}
It is a popular metric of evaluating a regression model. After observing the upper and lower control limits mentioned in section \ref{sec:data}, we found that the gap between upper and lower control limits is usually smaller if the ground truth has smaller absolute value. In other words, higher precision is usually needed if the ground truth has smaller absolute value. This motivates us to minimize the loss 
\begin{equation}
  \mathcal{L}_{RE}(\hat{y}, y)=\frac{|\hat{y}-y|}{\max(|y|,c)}
  \label{eq:REloss}
\end{equation}

In equation \ref{eq:REloss}, because we need to avoid the divisor being $0$ and avoid some samples being too influential, we use $\max(|y|,c)$ instead of $|y|$ as our denominator, where $c$ is a positive constant. And we choose $c=10$ in our experiments.

\subsubsection{Normalized $L_1$ Loss}
Relative Error Loss is simple and easy to use. It is a special case of weighted $L_1$ loss. 

To be more general, we can normalize each \emph{meas\_med} using two constants $b_1$ and $b_2$, where $b_1<b_2$, then train with $L_1$ loss. Ideally, $b_1$ and $b_2$ should be determined by some categorical variables within the measurement features. 

The formula to normalize the target $y$ is
\begin{equation}
  \Tilde{y}=\frac{y-b_1}{b_2-b_1}
  \label{eq:transform}
\end{equation}
Given a model output $\hat{\Tilde{y}}$, our Normalized $L_1$ loss function is 
\begin{equation}
  \mathcal{L}_{NL_1}(\hat{\Tilde{y}}, y)=|\hat{\Tilde{y}}-\Tilde{y}|
  \label{eq:NL1loss}
\end{equation}
During prediction, we can transform our model output back to the original scale by using 
\begin{equation}
  \hat{y}=\hat{\Tilde{y}}*(b_2-b_1)+b_1
  \label{eq:inv_transform}
\end{equation}

Intuitively, $b_1$ and $b_2$ should be similar to the lower and upper control limits mentioned in section \ref{subsec: lcl_ucl}, but in reality people may change the lower and upper control limits from time to time. Therefore, the choice of $b_1$ and $b_2$ for each wafer should be very careful. We need to specify $(b_1, b_2)$ so that it performs the normalization we want and at the same time do not confuse the model.

\section{Data}
\label{sec:data}
In this section, we introduce our datasets collected from Seagate factories and provide a way to preprocess this dataset.

\subsection{Sensor data and measurements data}
During manufacturing, a wafer goes through several processing stages including polishing, deposition, lithography, and etching. After each processing stage, the wafer is sent to metrology tools for quality control inspection. Multiple critical measurements are estimated, which are key quality indicators (KQIs). Accordingly, we get two mapping datasets, a sensor dataset and a metrology dataset. 

After dropping duplicated samples, the sensor dataset has a shape of (1301234, 144), and the metrology dataset has a shape of (4579232, 32). Both sensor and metrology datasets contain ID columns, i.e. processing ID and product ID. 
Each wafer can be uniquely identified by a unique pair of processing and product IDs. 

In the sensor dataset, except the \textit{hard} sensor types of data, it also contains the \textit{soft contextual} sensor data -- textual reports, such as textual information of the multi-stage manufacturing process, tools, processing modules, etc. These are referred to as categorical variables. There are 7 textual categorical columns in the sensor dataset and 8 categorical columns in the metrology dataset.

\begin{figure}[!t]
  \centering
  \includegraphics[width=2.5in]{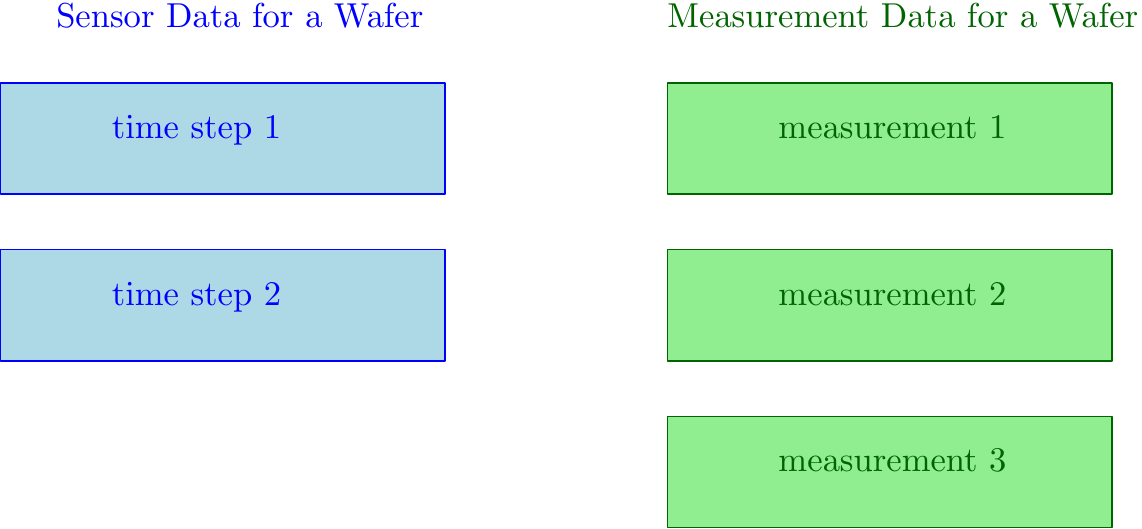}
  \caption{Sensor and measurement data.}
  \label{fig:sensor_and_meas}
\end{figure}

A wafer may have multiple sensor readings, we sort them in chronological order and call each row a \emph{(sensor) time step}. As shown in Figure \ref{fig:sensor_and_meas}, a wafer may have multiple metrology records as well, measured in different inspection stages using different measuring methods. 

\subsection{Monitor data and non-monitor data}

In the metrology process, a batch of wafers are fed into a tool which are a combination of product wafers and a monitor wafer. The actual measurements are taken on the monitor wafer. The other wafers' measurements data inherit the metrology measurements from that monitor wafer's measurements data. However, the specifications that define the \emph{pass/fail} for wafer are based on the product wafer.

We thus use monitor data for training the regression model and use non-monitor (product wafer) data for checking the \emph{pass/fail} results of wafers.

\begin{figure}[!t]
  \centering
  \includegraphics[width=2.5in]{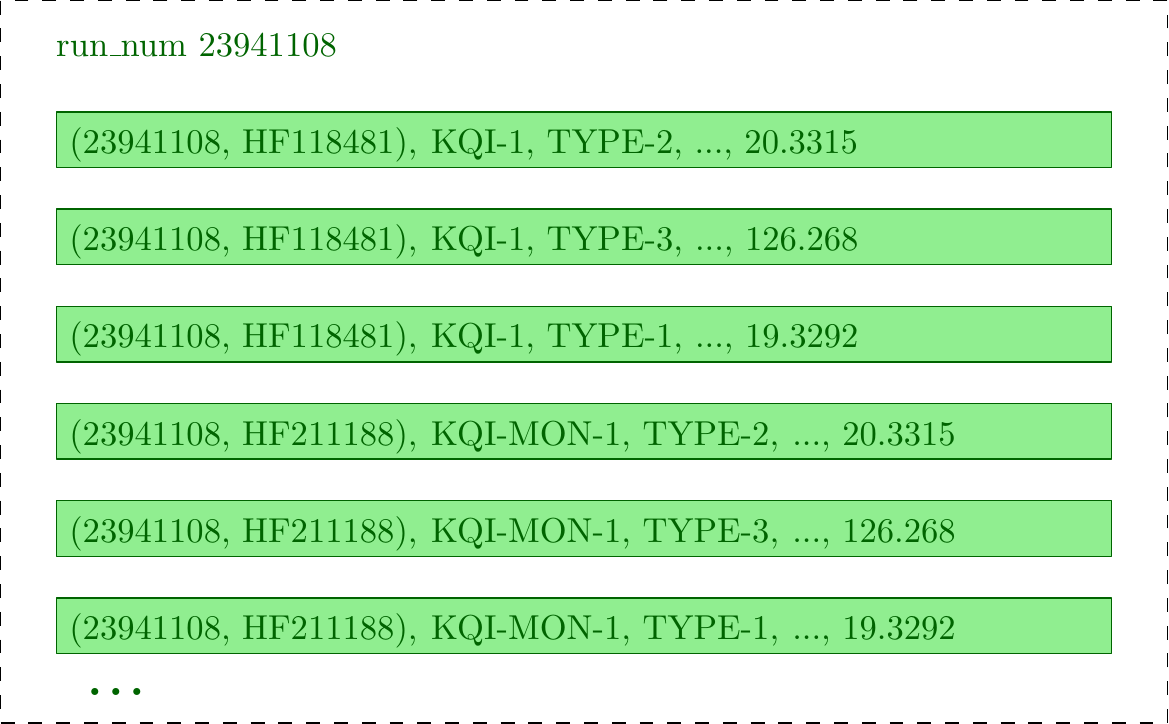}
  \caption{An exemplary illustration of Data.}
  \label{fig:meas_mon}
\end{figure}

An example can be seen in Figure \ref{fig:meas_mon}. They all belong to processing id 23941108, but the KQIs of some rows are KQI-1, while some other rows have KQI-MON-1. There are two samples with measuring method TYPE-1, and their measuring results are both 19.3292. This is because engineers assume that these two samples should be the same. Only the sample with KQI-1 has a real metrology measurement, so the sample with KQI-MON-1 inherits the results from the sample with KQI-1.

After our data preprocessing, there are 2601336 rows in the monitor data and 1845720 rows in the non-monitor data.

\subsection{Lower and upper control limits}
\label{subsec: lcl_ucl}
There are two datasets that are related to lower and upper control limits. One dataset provides two numbers \emph{lcl} and \emph{ucl} for each type of wafer. The other dataset is the measurements data, which provides two numbers \emph{targ\_min} and \emph{targ\_max} in some rows. Both (lcl, ucl) and (targ\_min, targ\_max) are meant to be the lower and upper control limits.

Because \emph{targ\_min} and \emph{targ\_max} are usually more accurate than \emph{lcl} and \emph{ucl}, if \emph{targ\_min} and \emph{targ\_max} exist, we assume them to be the lower and upper control limits. Otherwise we use \emph{lcl} and \emph{ucl} as the lower and upper control limits.

\subsection{Pass/fail labels}
\label{subsec:pflabels}
The \emph{passfail} column in measurements data contains different types of labels. Since we only predict \emph{meas\_med}, we only focus on predicting three types of labels: PASS, FAIL\_AVG\_HI and FAIL\_AVG\_LOW. FAIL\_AVG\_HI means the \emph{meas\_med} is larger than the upper control limit, and FAIL\_AVG\_LOW means the  \emph{meas\_med} is less than the lower control limit.


\subsection{Data preprocessing}

\subsubsection{Before joining sensor and measurements data}
The data is split into training, validation and test set by ratio 7:2:1. 
Due to the constraint of storage and computing resources, we only focus on 33 most common \emph{(KQI, TYPE)} of wafers.

In order to deal with date and time information in the sensor data, we convert it into two features: time in the current day and date in the current year.

Because different columns have different scales, we normalize each column in the sensor training set to be between 0 and 1. Then use the same min-max scaler to transform the validation and test set. 

There are lots of missing values in the sensor readings. we decide to fill in these missing values with the medians in the columns they belong to. But we drop those columns which contains only missing values or has no variability.

Those samples with \emph{meas\_med} outside the range [-1,1000] are dropped in the training set because they are likely to be measurement errors. The dropped samples only represent less than 1\% of the whole dataset.  

One-Hot Encoding are performed on the 5 categorical columns in the sensor data and 5 categorical columns in the measurements data. After that, we have 267 and 552 features in each sensor row and measurement row.

\subsubsection{Join sensor and measurements data}

During model training, for every sample we need to use both the sensor data and measurements data, so it would be helpful if we can join them before model training. Note that we only need non-monitor measurements data in our regression task, we only need to join sensor data and non-monitor measurements data. 

\begin{figure}[!t]
  \centering
  \includegraphics[width=2.5in]{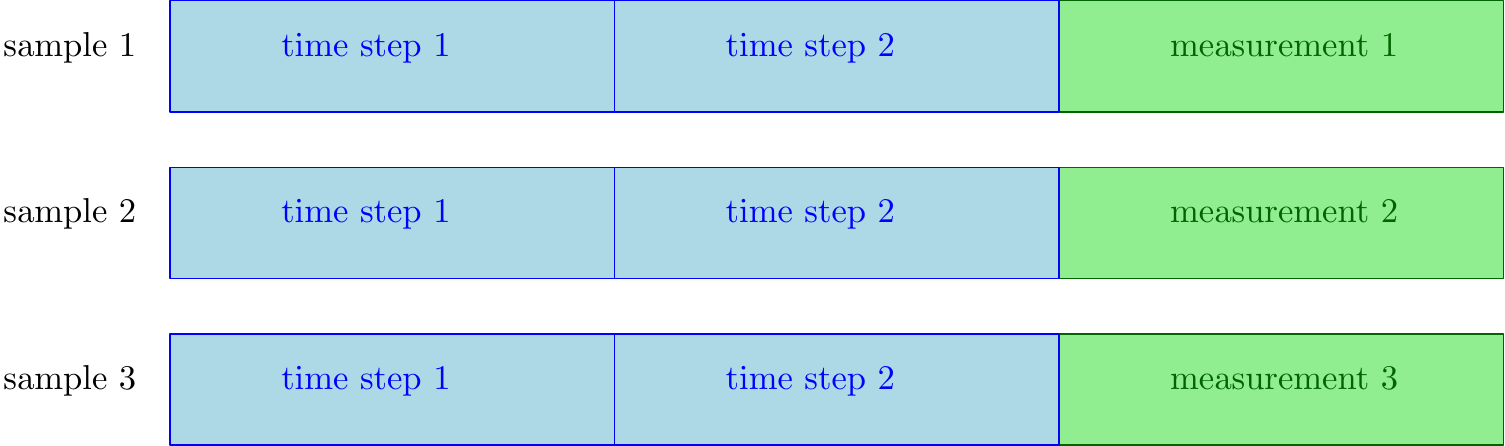}
  \caption{Join sensor data and measurement data.}
  \label{fig:join}
\end{figure}

An example can be found in Figure \ref{fig:join}. Let's consider a wafer which has two time steps and three measurements. We concatenate these two time steps in a row and join it with every one of the three measurements. So we end up getting three samples. Each sample consists of a row which has 267*2+552=1086 features. We apply this \emph{join} operation to wafers which have two time steps and stack the samples. 

However, if another wafer has five time steps, then after this operation it will have 267*5+552=1887 features. So we need to store these 1887-feature samples in another file.

In the training set and validation set, the number of time steps of wafers can be 1, 2, 3, 4, 5, 6, 7, 8 or 9.  In the test set, we only see 1, 2, 3, 4, 5 and 8 time steps. Most wafers have two or three time steps.

When we load our joined data and do the training or evaluation, each batch should contain wafers which have the same number of time steps. And it's not difficult to reshape the sensor features back into sensor time steps. 

\section{Results}
\label{sec:results}
\subsection{Evaluation metrics}
\subsubsection{Error Grouping}
When $|y|$ is small, \emph{Relative Error} $\eta$ can easily get very large and can be meaningless, but \emph{Absolute Error} $\epsilon=|\hat{y}-y|$ may still work. So we propose the following grouping criteria for prediction errors:

\begin{itemize}
\item Group 1: $\eta<1\%$ or $\epsilon<0.1$.

\item Group 2: $\eta<5\%$ or $\epsilon<0.5$, and not in Group 1.

\item Group 3: $\eta<10\%$ or $\epsilon<1$, and not in Group 1 or 2.

\item Group 4: $\eta<50\%$ or $\epsilon<5$, and not in Group 1 to 3.

\item Group 5: $\eta<100\%$ or $\epsilon<10$, and not in Group 1 to 4.

\item Group 6: not in Group 1 to 5.
\end{itemize}

Apparently, a group with smaller index is a better group. We report grouping results in our experiments, and define `decent predictions' as those predictions in group 1 and group 2. 

\subsubsection{Recall and false positive rate}
\label{subsubsec:recall_fpr}
The \emph{passfail} column and human inspection results both give us insights into the pass or fail status of wafers, as introduced in section \ref{subsec:pflabels}. In our evaluation, we use the monitor data in measurements data and define a fail wafer as a wafer which satisfies all the following conditions:
\begin{itemize}
\item The value in its \emph{passfail} column is FAIL\_AVG\_HI or FAIL\_AVG\_LOW.

\item The result of human inspection is REWORK or SCRAP.

\item Its \emph{meas\_med} is either larger than the upper control limit or smaller than the lower control limit.
\end{itemize}

The pass and fail labels in our dataset is very imbalanced. We only have a small number of fail wafers. Fail wafers are regarded as positive samples. We mostly care about the recall rate of fail wafers and the false positive rate.

\subsection{Compare two losses}
In this subsection, we compare the Relative Error Loss with the Normalized L1 Loss. We group the data by \emph{(KQI,TYPE,stage)} and call each group a normalization group. In the Normalized L1 Loss, we normalize \emph{meas\_med} within each \emph{(KQI,TYPE,stage)} normalization group, then train with L1 loss like what we mentioned in section \ref{subsec:losses}. Although there are two other categorical variables: \emph{equipid} and \emph{prod}, it seems that they are not as important as \emph{(KQI,TYPE,stage)}. Another reason we did not use \emph{equipid} and \emph{prod} is that, we do not want to have too many normalization groups in the normalization. Using \emph{(KQI,TYPE,stage)}, we have 74 normalization groups in the entire training set, which we think is acceptable compared to our sample size. 

Within each \emph{(KQI,TYPE,stage)} normalization group, the constants pair $(b_1,b_2)$ is chosen to be the pair of lower and upper control limits which has the smallest $b_2-b_1$. 

\subsubsection{LSTM-small model}

Our model architecture is explained in section \ref{subsec:models}. If we want to train on only one \emph{(KQI, TYPE)}, we can use relatively less parameters to define a LSTM-small model. Later when we train on all data, we define a LSTM-large model with more parameters. 

In the LSTM-small model, our input size and hidden size of LSTM are both 128. In the MLP regressor, we have two hidden layers, each with 256 hidden units. In total there are approximately 0.4 million parameters in the LSTM-small model. Our activation function in the MLP regressor is ReLU. We stop the training if we observe no improvement of the validation loss for more than 10 continuous epochs. Our optimizer is Adam with learning rate 0.0001. After some hyperparameter tuning, we set the batch size to be 16. We keep the model with the best validation loss.

\begin{table}[!t]
  \caption{LSTM-small model on (KQI-1, TYPE-1)\label{tab:tab1}}
  \centering
  \begin{tabular}{|c||c||c|}
    \hline
    \textbf{Model} & \textbf{Loss}& \textbf{Relative Error Grouping}\\
    \hline
    LSTM-small                       & RE & [2554, 2859,  257,   68,   10,    1]    \\
    LSTM-small                      & NL1  & [2544, 2726,  319,  100,   36,   24]  \\
    \hline
  \end{tabular}
\end{table}

Table \ref{tab:tab1} summarizes the test set results when we train our LSTM-small model on \emph{KQI} KQI-1 and \emph{TYPE} TYPE-1. We choose this \emph{(KQI, TYPE)} because it is one of the most common \emph{(KQI, TYPE)}s. In the table, \emph{RE} means the training loss is the Relative Error Loss. And \emph{NL1} means the Normalized $L_1$ Loss. Each row in the last column contains a list of six numbers, representing the number of test samples in each group under our grouping criteria. 

According to our experiments, Relative Error Loss gives us better results. Since we have already defined predictions within group 1 and group 2 as decent predictions, we have $5413/5749\approx 94.16\%$ decent predictions using RE loss, while we only have $5270/5749\approx 91.67\%$ decent predictions using NL1 loss. Considering that our ground truth can vary from $0$ to as much as $900$, $94.16\%$ of the predictions are decent predictions is a pretty good result.

\subsubsection{LSTM-large model}
Because we have 33 different \emph{(KQI, TYPE)}s in our dataset, it may be interesting to train all \emph{(KQI, TYPE)}s jointly and see if we can improve the results. 

In order to do this, we use a much larger model called LSTM-large model. Its input size and hidden size of LSTM are both 1024. Its MLP regressor still has 2 hidden layers but each hidden layer has 2048 units. LSTM-large model has approximately 16 million parameters in total. We stll use batch size 16 and Adam optimizer with learning rate 0.0001. 

\begin{table}[!t]
  \caption{LSTM-large model on all data
  \label{tab:tab2}}
  \centering
  \begin{tabular}{|c||c||c|}
    \hline
    \textbf{Model} &\textbf{Loss}&\textbf{Relative Error Grouping}\\
    \hline
    LSTM-large                       & RE & [10763, 12311,  2122,  2899,    32,     7]     \\
    LSTM-large                      & NL1  & [12812,  9606,  2071,  3458,    98,    89]     \\
    \hline
  \end{tabular}
  
\end{table}

The results are displayed in Table \ref{tab:tab2}. By Relative Error Loss $23074/28134\approx 82.01\%$ predictions are decent predictions. This is worse than the $94.16\%$ for LSTM-small model on (KQI-1, TYPE-1), perhaps because (KQI-1, TYPE-1) is a relatively easy \emph{(KQI, TYPE)}.

By Normalized L1 Loss, we have $22418/28134\approx 79.68\%$ decent predictions, which is still worse than the result by Relative Error Loss. However, if we only care about the number of predictions within group 1, model trained with Normalized L1 Loss has 12812 samples in group 1, which is better than the 10763 samples by Relative Error Loss.

\begin{table}[!t]
  \caption{LSTM-large model on (KQI-1, TYPE-1)
  \label{tab:tab3}}
  \centering
  \begin{tabular}{|c||c||c|}
    \hline \textbf{Model} & \textbf{Loss}&\textbf{Relative Error Grouping}\\
    \hline
    LSTM-large                       & RE & [2380, 3039,  231,   78,   19,    2]     \\
    LSTM-large                      & NL1  & [2626, 2791,  210,   55,   42,   25]     \\
    \hline
  \end{tabular}
\end{table}

Another experiment we find interesting is using the same LSTM-large model trained on all 33 \emph{(KQI, TYPE)}s but test only on (KQI-1, TYPE-1). We have $5419/5749\approx 94.26\%$ decent predictions by \emph{RE} Loss and $5417/5749\approx 94.23\%$ decent predictions by \emph{NL1} Loss, see Table \ref{tab:tab3}. Both $94.26\%$ and $94.23\%$ are better than their counterparts in LSTM-small model, which makes sense because we have much more training data when we train all 33 \emph{(KQI, TYPE)}s jointly. 

Comparing the percentage of decent predictions, we think Relative Error Loss is slightly better than Normalized L1 Loss in our settings. But Normalized L1 Loss has its own advantages. For example, after the same normalization, we can easily replace the L1 loss by L2 loss or Huber loss so Normalized L1 Loss is more general. Besides, the Normalized L1 Loss utilizes the lower and upper control limits while the Relative Error Loss does not. This may be beneficial if accurate lower and upper control limits are available.

\subsection{Pass/fail evaluation}
The definition of a fail wafer can be found in section \ref{subsubsec:recall_fpr}. Under that definition, we only have 162 fail wafers among 550239 training samples. 

Using the LSTM-large model trained on all \emph{(KQI, TYPE)}, and we predict a wafer to be a fail wafer if and only if the predicted meas\_med $\hat{y}$ is outside the interval $(b_1^*, b_2^*)$, we get recall rate as $0.3580$ and false positive rate as $0.01434$. The $b_1^*$ and $b_2^*$ are the lower and upper control limits introduced in section \ref{subsec: lcl_ucl}.

However, people may want to sacrifice false positive rate for a better recall rate, so we apply the following trick here: Consider a constant $0<f<0.5$, and $r = b_2^*-b_1^*$, we can predict the wafer as a fail wafer if and only if 
\begin{equation}
  \hat{y} \notin (b_1^*+f*r, b_2^*-f*r)
  \label{eq:tradeoff}
\end{equation}
When $f$ increases, the recall rate and false positive rate of the given model should both increase because we predict more and more wafers to be fail wafers.

\begin{table}[!t]
  \caption{Recall and False Positive Rate}
  \label{tab:recall_fpr}
  \centering
  \begin{tabular}{|c||c||c|}
    \hline \textbf{$f$} & \textbf{Recall}& \textbf{False Positive Rate }\\
    \hline
    0.0 & 0.3580 &  0.01434    \\
    0.1 & 0.4506  &   0.07219   \\
    0.2 & 0.4938 &  0.1084   \\
    0.3 & 0.6975 &  0.2075    \\
    0.35  &  0.8086 &  0.2981    \\
    0.4  & 0.8395 & 0.4321     \\
    \hline
  \end{tabular}
\end{table}

The results can be found in Table \ref{tab:recall_fpr}, where a $f=0.35$ gives us a recall rate above $0.8$. At this time the false positive rate is $0.2981$, which is pretty high. But our model is still useful because if we think 80\% recall rate is good enough, then Table \ref{tab:recall_fpr} shows that we now only need to apply metrology tools on only approximately 30\% of the wafers where $\hat{y}\notin (b_1^*+0.35r, b_2^*-0.35r)$.

Due the fact that there seems to have some wrong fail wafers in our dataset, the actual recall rate should be higher. And our results will be more accurate if we collect more test data in the future.

\section{Discussions and Conclusion}
\label{sec:discussion_conclusion}
For our soft sensing regression problem, we have designed two different loss functions, i.e. Relative Error Loss and Normalized L1 Loss. We have also designed a model based on LSTM. From the evaluation perspective, we have established a few criteria on evaluating a model in soft sensing regression.

Our results shows that using deep learning to predict measurements based on sensor time steps is not only possible but also promising. With deep learning, we can help the factories rely less on metrology tools and save a lot of time and energies. 

In order to improve the model, we think the most important thing is to remove outliers in the data. We have observed some measurements in the data that are apparently wrong. These wrong data can cause damages in both model training and model evaluation. After removing outliers, it may be helpful to assign more weights to those fail wafers in the model training because we care more about the recall rate. 

Another interesting direction is to do transfer learning on different soft sensing datasets which come from different tool sets at Seagate, and improve our model's generalization ability.

\section*{Acknowledgments}
The authors would like to thank Seagate Technology for providing the data and the computational resources. We should especially thank the engineers in Seagate Technology who helped us a lot in understanding the data.

\bibliographystyle{IEEEtran}
\bibliography{mybib}

\end{document}